\documentclass[10pt,twocolumn,letterpaper]{article}

\usepackage{cvpr}              

\usepackage{graphicx}
\graphicspath{{figs/}} 
\DeclareGraphicsExtensions{.pdf,.png,.jpg} 

\definecolor{cvprblue}{rgb}{0.21,0.49,0.74}
\usepackage[pagebackref,breaklinks,colorlinks,allcolors=cvprblue]{hyperref}

\def\paperID{****} 
\def\confName{CVPR}
\def\confYear{2026}

\title{DSAA: Dual-Stage Attribute Activation \\
for Fine-grained Open Vocabulary Detection}

\author{
Donghong Jiang$^{1*}$, Endian Lin$^{1*}$, Hanqing Liu$^{1}$, Mingjie Liu$^{1}$,
Luoping Cui$^{1}$, Zhao Yang$^{2}$, Chuang Zhu$^{1,3\dagger}$\\[6pt]
$^{1}$  Beijing University of Posts and Telecommunications
$^{2}$  Beijing E-Hualu Information Technology Co., Ltd.\\
$^{3}$ State Key Laboratory of General Artificial Intelligence, BIGAI, Beijing, China\\[4pt]
{\tt\small \{donghongjiang,ledgogo,hanqingliu,LMJ,lpcui,czhu\}@bupt.edu.cn, zhaoy01@ehualu.com}
}
\begin{document}
\maketitle
\begingroup
\renewcommand\thefootnote{}
\footnotetext{* Equal contribution.}
\footnotetext{\textdagger\ Corresponding author.}
\endgroup
\begin{abstract}
\textit{Open-Vocabulary Object Detection (OVD) models break the limitations of closed-set detection, enabling the identification of unseen categories through natural language prompts. However, they exhibit notable limitations in fine-grained detection tasks involving attributes like color, material, and texture. We attribute this performance bottleneck in OVD models to a core issue: when category signals dominate, OVD models tend to marginalize attribute information during inference. This leads to incorrect binding between attributes and target objects. To address this, we propose the Dual-Stage Attribute Activation (DSAA) framework, which enhances fine-grained detection capabilities by strengthening attribute semantics at two critical stages. In the text embedding stage, we employ Attribute Prefix Adapter (APA) module to generate attribute prefixes that inject explicit attribute priors. To further amplify the influence of these attributes, our Key/Value (K/V) Modulator module then intervenes during the BERT encoding phase, selectively enhancing the Key and Value vectors of the corresponding attribute tokens. In addition, we introduce an attribute-aware contrastive loss to improve discrimination among same-category instances with different attributes during training. Experimental results on the FG-OVD benchmark demonstrate the effectiveness of our method across various mainstream open-vocabulary models.}
\end{abstract}    
\section{Introduction}
\label{sec:intro}

Open-Vocabulary Object Detection (OVD) \cite{feng2022promptdet,li2024cliff,gu2021open,cheng2024yoloworld}  has emerged as a pivotal paradigm in computer vision, breaking the shackles of closed-set detection \cite{ren2016faster,redmon2016you,zhu2020deformable} by enabling models to recognize arbitrary categories through natural language prompts. This capability endows it with a wide range of application scenarios, such as autonomous driving \cite{ma2022rethinking,xia2024openad,ilyas2024potential}, human-computer interaction\cite{levi2023object,liu2024objectfinder} and robotics \cite{huang2023voxposer, liang2023code,van2024open,zhang2025ovgrasp,chikhalikar2025open}.
Fueled by large-scale vision-language pre-training, advanced open-vocabulary detectors like Grounding DINO \cite{liu2024grounding} have demonstrated remarkable zero-shot capabilities on standard benchmarks such as COCO \cite{lin2014coco} and LVIS \cite{gupta2019lvis}. 

\begin{figure}[t]
    \centering
    \includegraphics[width=\linewidth]{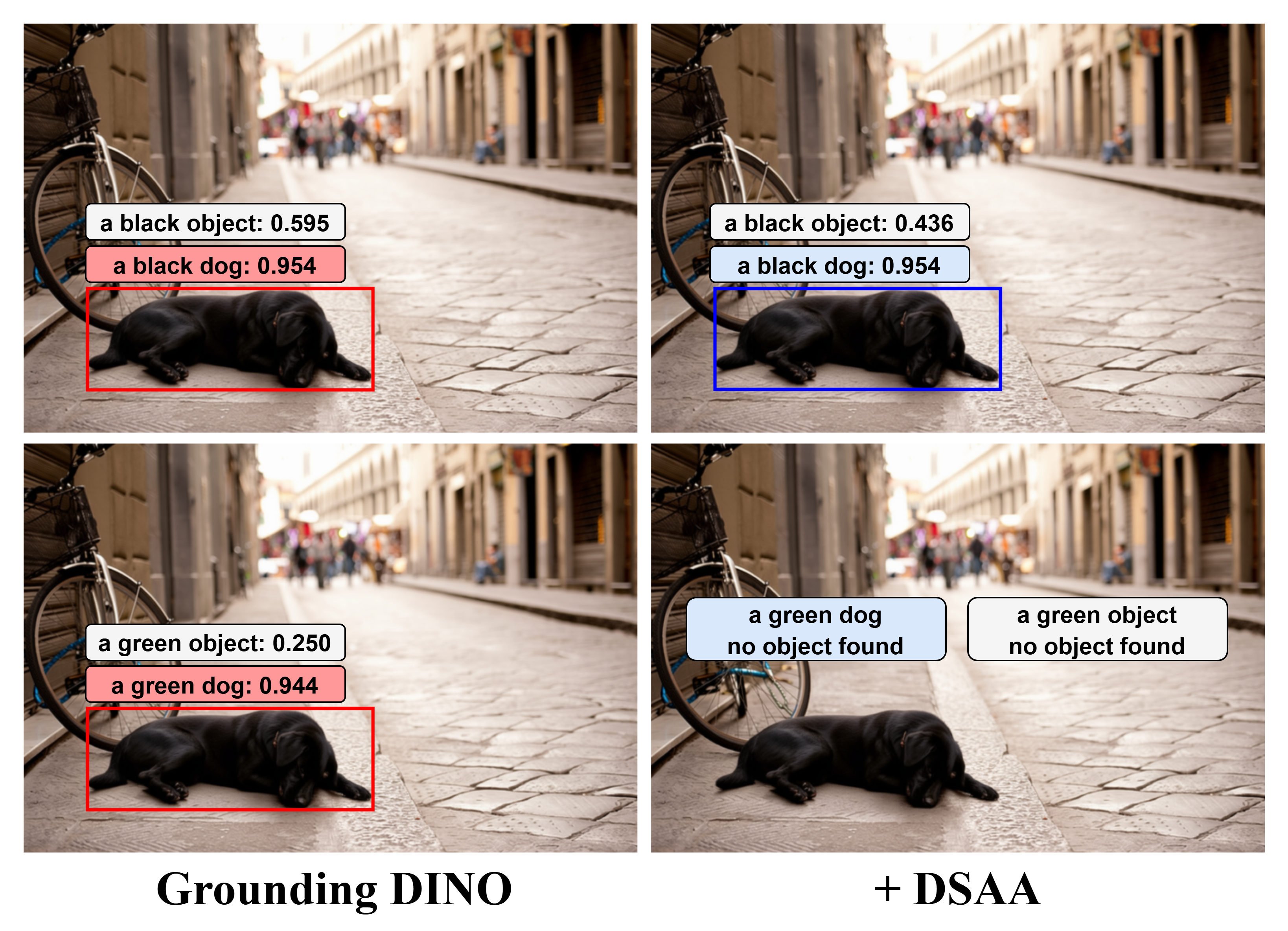}
    \caption{
    \textbf{Motivating example comparing Grounding DINO and DSAA on attribute-sensitive prompts.}
    The baseline model (left) confuses attribute semantics, assigning high confidence to invalid compositions such as \textit{``a green dog''}, 
    while DSAA (right) correctly rejects mismatched queries and preserves consistent attribute–object binding. 
    }
    \label{fig:motivation}
\end{figure}

However, a critical challenge remains insufficiently explored: enabling these models to perform reliably in fine-grained scenarios where attribute understanding is important.
When real-world queries contain intra-class attribute modifiers, such as ``a red leather chair'' or ``a black dog with a red hat'', existing OVD models demonstrate considerable limitations. To systematically evaluate this capability gap, the Fine-grained Open Vocabulary Detection (FG-OVD) benchmark~\cite{bianchi2024devil} is introduced, which specifically challenges intra-class attribute discrimination. The benchmark requires models to accurately localize targets by resolving attribute semantics within compositional phrases. Empirical evaluations reveal that even advanced  OVD models like Grounding DINO suffer marked performance declines, highlighting a fundamental limitation in fine-grained understanding across current OVD architectures.

We contend that this challenge does not stem from the attribute knowledge deficit, but rather from a failure of activation — the model knows the attributes but fails to activate them when strong category priors dominate. While extensive pre-training endows OVD models with rich attribute representations, these semantics are systematically suppressed when strong category priors dominate, revealing a fundamental yet largely overlooked limitation of current OVD systems. This limitation is clearly exemplified by the behavior of Grounding DINO under attribute-sensitive prompts. As illustrated in \cref{fig:motivation}, when presented with an image of a black dog, the model confidently detects it as ``green dog'' but refuses to detect it as ``green object'', while both ``black dog'' and ``black object'' prompts succeed.
This asymmetry indicates that the model retains attribute sensitivity for category-neutral terms such as ``object'', but loses attribute sensitivity once the attribute is paired with a semantically strong category like ``dog''. In real-world queries, where attributes typically co-occur with specific object categories, the dominant category signal suppresses attribute cues.
This results in incorrect attribute–object binding and ultimately degrading fine-grained discrimination.

To address this, we introduce Dual-Stage Attribute Activation (DSAA), a non-invasive framework that systematically activates and amplifies attribute information at two critical stages.  As shown in \cref{fig:motivation}, DSAA effectively resolves the attribute confusion issue observed in baseline models, enabling correct rejection of mismatched attribute–category compositions. A key factor behind this improvement is that DSAA explicitly targets the stages where attribute semantics are most likely to be suppressed, providing a principled and elegant solution to this overlooked problem. First, in the text embedding stage, an Attribute Prefix Adapter (APA) converts extracted keywords into attribute prefix tokens and injects them into text embeddings as explicit priors, establishing clear semantic anchors for fine-grained distinctions. Subsequently, in the text encoding stage, a Key/Value (K/V) Modulator intervenes in early BERT \cite{devlin2019bert} layers to selectively modulate the Key and Value vectors of corresponding attribute tokens, ensuring their semantics are robustly preserved and strengthened throughout the attention and feature aggregation processes. With only a minimal increase in parameters, DSAA can be seamlessly incorporated into various BERT-based open-vocabulary detection models, significantly activating their fine-grained detection capabilities.

In summary, our main contributions are as follows:
\begin{itemize}
    \item We propose \textbf{DSAA}, a non-invasive framework that mitigates attribute marginalization in open-vocabulary detectors by reinforcing attribute representations at both text embedding and encoding stages.
    \item We design two complementary modules in DSAA: (1) the \textbf{APA}, which injects explicit attribute priors at the text embedding stage, and
    (2) the \textbf{K/V Modulator}, which preserves and amplifies attribute semantics in early encoder layers. In addition, we introduce an \textbf{attribute-aware contrastive loss} to improve discrimination between same-category instances with different attributes.
    \item Extensive experiments on the challenging FG-OVD benchmark confirm the effectiveness and broad applicability of DSAA, which achieves a new state of the art with an average improvement of +20.5 mAP on Grounding DINO.
\end{itemize}
\section{Related Work}

\noindent \textbf{Open-Vocabulary Object Detection.}  
Open-vocabulary object detection (OVD) aims to detect both seen and unseen categories by leveraging the semantic knowledge of large vision–language models (VLMs)~\cite{cheng2024yoloworld,li2022glip,minderer2022simple,shi2023edadet,yao2023detclipv2,yao2024detclipv3,zhang2022glipv2}.  
Early works such as ViLD~\cite{gu2021open} and OV-DETR~\cite{zang2022open} adapt CLIP~\cite{radford2021learning} representations for zero-shot detection via distillation and contrastive alignment.  
Later methods including GLIP~\cite{li2022glip}, OWL-ViT~\cite{minderer2022simple} and Grounding DINO~\cite{liu2024grounding} reformulate detection as a grounding task that aligns region proposals with text queries.  
RegionCLIP~\cite{zhong2022regionclip} and DetCLIP~\cite{yao2023detclipv2} further strengthen region–text alignment through large-scale pretraining.  
Although these approaches achieve strong category-level generalization, they mainly emphasize category correspondence and are not explicitly designed for fine-grained attribute discrimination.

\noindent \textbf{Fine-Grained Understanding in OVD.}
Fine-grained understanding in OVD requires models to reason over both category and attribute cues (e.g., distinguishing ``brown dog'' from ``black dog'').
Recent studies have revealed that mainstream OVD models exhibit substantial performance degradation on fine-grained detection tasks involving subtle attributes~\cite{bianchi2024devil}.   
Several works attempt to enhance textual semantics for better fine-grained understanding. DesCo~\cite{li2023desco} enriches text representations with LLM-generated descriptive captions, LaMI-DETR~\cite{du2024lami} improves category descriptions using GPT-3.5, and SHiNe~\cite{liu2024shine} updates OVD classifiers with hierarchy-aware sentences. While these methods improve semantic richness, they are not specifically designed for attribute-conditioned fine-grained open-vocabulary detection. 
HA-FGOVD~\cite{ma2025ha} is the most closely related work to ours. It assumes that pretrained OVD models already contain attribute-sensitive features in their frozen text representations, and improves fine-grained detection by extracting attribute-specific features through token masking and explicitly linearly composing them with global text features. In essence, HA-FGOVD operates as an external feature recomposition approach on top of frozen encoder outputs. In contrast, DSAA is motivated by a different diagnosis: attribute semantics are not merely under-emphasized at the output level, but are progressively suppressed inside the text encoder by dominant category signals. Therefore, instead of recombining text features after encoding, DSAA directly intervenes in the encoder itself. Specifically, APA injects attribute-specific prefixes at the embedding stage to establish semantic anchors, the K/V Modulator reinforces attribute signals during early contextual encoding, and the attribute-aware contrastive loss explicitly improves separability between same-category instances with different attributes. As a result, while HA-FGOVD mainly highlights existing attribute features after encoding, DSAA focuses on the internal activation, preservation, and propagation of attribute semantics throughout the encoding process.

\section{Method}

\begin{figure}[t]
    \centering
    \includegraphics[width=\linewidth]{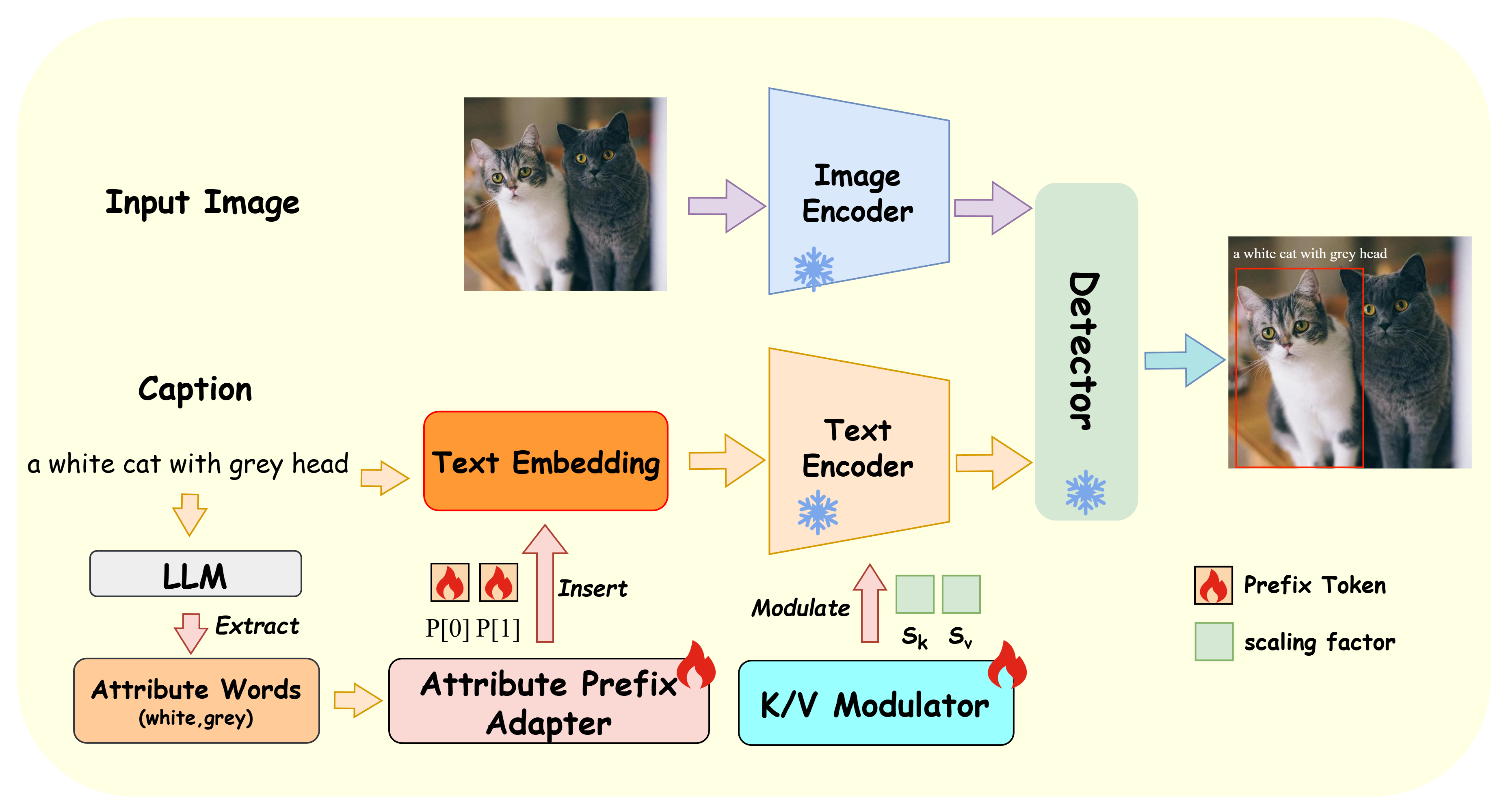}
    \caption{
    \textbf{Inference pipeline with the proposed Dual-Stage Attribute Activation (DSAA).}
    DSAA activates fine-grained attribute semantics in two stages:
    (1) An Attribute Prefix Adapter injects explicit attribute priors into the text embedding space, 
    and (2) a K/V Modulator selectively amplifies attribute tokens within early text encoder layers. 
    }
    \label{fig:dsaa_framework}
\end{figure}

\label{sec:method}
\begin{figure*}[t]
  \centering
  \includegraphics[width=\linewidth]{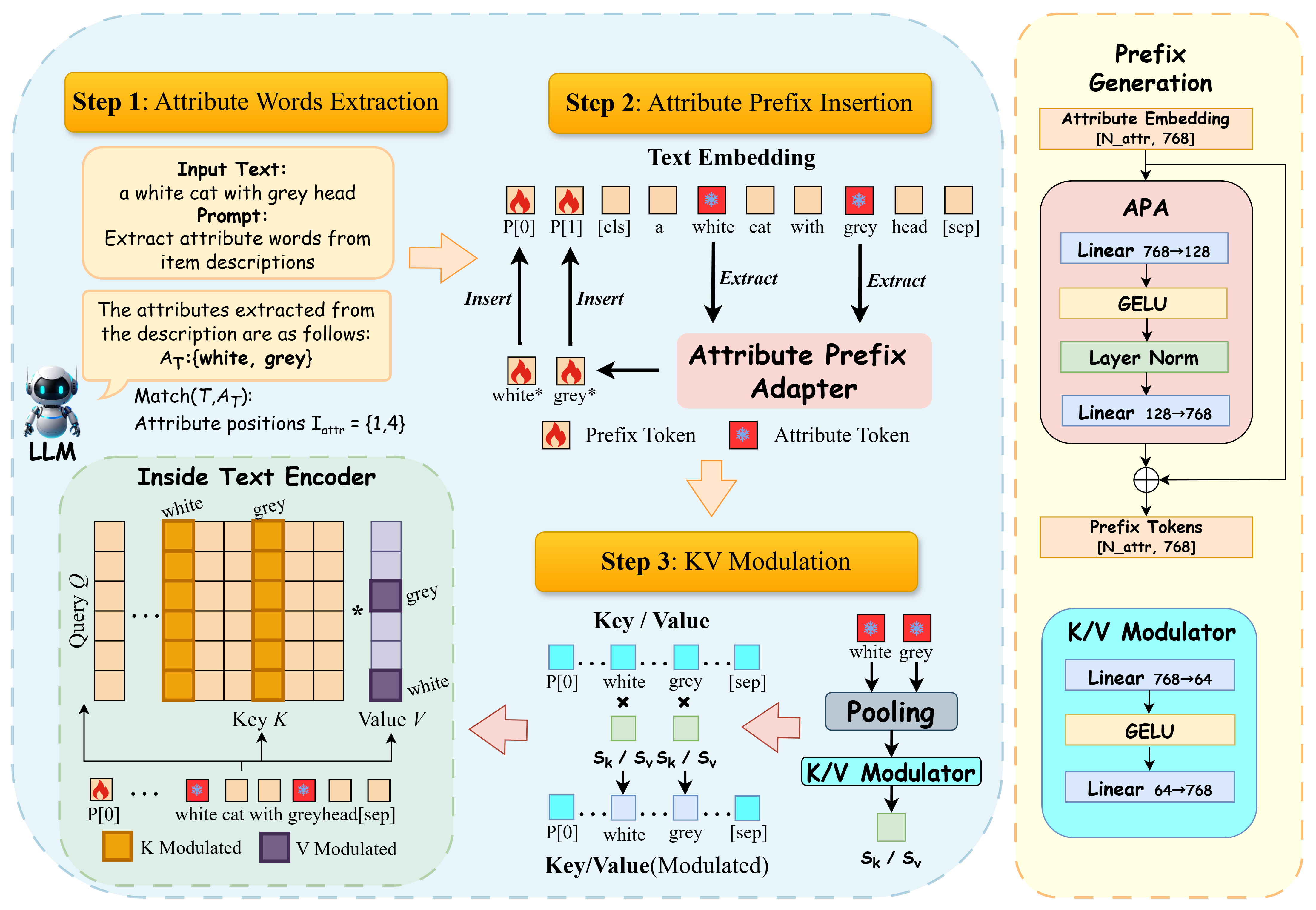} 
  \caption{
  \textbf{Overview of the proposed Dual-Stage Attribute Activation (DSAA) framework.}
    The overall workflow consists of three main components:
    (1) Attribute Words Extraction: an LLM identifies attribute words and their token positions from input text;
    (2) Attribute Prefix Insertion: extracted attributes are converted by Attribute Prefix Adapter into prefix tokens and inserted into text embeddings as attribute semantic anchor;
    (3) KV Modulation: in early BERT layers, the Key and Value channels of attribute tokens are dynamically scaled to enhance attention and preserve attribute semantics.
    The left panel (\emph{Inside Text Encoder}) illustrates the behavior of the text encoder after applying DSAA.
    The right panel illustrates the prefix generation process and the internal structure of the K/V Modulator.}
  \label{fig:DSAA_overview}
  \vspace{-0.75em}
\end{figure*}

We propose DSAA, a novel and non-invasive framework designed to mitigate the problem of attribute marginalization in existing open-vocabulary detection (OVD) models, which limits their performance on fine-grained OVD tasks. 
The inference pipeline with DSAA is shown in \cref{fig:dsaa_framework}.
DSAA consists of two lightweight architectural modules, namely the Attribute Prefix Adapter (APA) and the K/V Modulator, together with an attribute-aware contrastive loss for fine-grained supervision. 





\subsection{Attribute Prefix Insertion}

Within the DSAA framework, the Attribute Prefix Adapter (APA) serves as the first-stage module responsible for activating attribute semantics at the input of the text encoder.
At this stage, APA converts extracted attribute keywords into attribute prefix tokens and injects them into the text embeddings as explicit attribute priors.
These prefix tokens serve as semantic anchors that preserve and highlight attribute information, making it explicitly accessible to subsequent transformer layers and preventing it from being suppressed by dominant category semantics.

Before injection, we first identify the attribute words or phrases contained in each input caption $T = (t_1, t_2, \ldots, t_L)$. To achieve this, we employ the Qwen3~\cite{yang2025qwen3} language model under an instruction prompt $P$ to extract the attribute set:
\begin{equation}
\mathcal{A} = \text{Extract}_{\text{LLM}}(P, T).
\end{equation}
After tokenization, token-level alignment is performed to locate the corresponding positions of these attributes in the text sequence:
\begin{equation}
\mathcal{I}_{\text{attr}} = \text{Match}(\text{Tokenizer}(T), \text{Tokenizer}(\mathcal{A})),
\end{equation}
where $\mathcal{I}_{\text{attr}} \subseteq \{1, \ldots, L\}$ denotes the index set of attribute tokens. This preprocessing step aligns the word-level outputs of the LLM with the token-level representation space of the text encoder, ensuring accurate correspondence between extracted attributes and their token positions for subsequent prefix injection by APA.

For each aligned attribute token position $i \in \mathcal{I}_{\text{attr}}$, we obtain its embedding vector $\mathbf{a}_i \in \mathbb{R}^D$ at the embedding stage and feed it into APA for prefix generation. As illustrated in \cref{fig:DSAA_overview}, APA is a lightweight residual adapter composed of two linear layers with a bottleneck dimension $d$, a GELU activation, and a Layer Normalization layer. Given an attribute embedding $\mathbf{a}_i \in \mathbb{R}^D$, the generated prefix token is computed as
\begin{equation}
\mathbf{P}_i=
\frac{
\mathbf{a}_i + W_2\,\mathrm{LN}\!\left(\sigma(W_1\mathbf{a}_i)\right)
}{
\left\|
\mathbf{a}_i + W_2\,\mathrm{LN}\!\left(\sigma(W_1\mathbf{a}_i)\right)
\right\|_2
}
\cdot \|\mathbf{a}_i\|_2,
\end{equation}
where $W_1 \in \mathbb{R}^{d \times D}$ and $W_2 \in \mathbb{R}^{D \times d}$ are trainable projection matrices, $\sigma(\cdot)$ denotes the GELU activation, and $\mathrm{LN}(\cdot)$ represents Layer Normalization. The final rescaling keeps the generated prefix at a scale comparable to the original attribute embedding.

Let $k = |\mathcal{I}_{\text{attr}}|$ denote the number of aligned attribute tokens. The resulting prefix vectors $\mathbf{P}=[\mathbf{P}_{1:k}]$ are concatenated before the original token embeddings to form the augmented input $[\mathbf{P}_{1:k}, \mathbf{e}_{1:L}]$.

This input-level semantic activation establishes explicit attribute anchors at the earliest layer, enabling the model to capture and reinforce fine-grained attribute information from the beginning of encoding without modifying the detector backbone.

\subsection{Attribute KV Modulation}

While prefix tokens introduce global attribute priors, they are insufficient to ensure consistent propagation of attribute semantics throughout the encoder.
To further reinforce attribute semantics at the attention level, we introduce Key/Value (K/V) Modulator as our second-stage module, which dynamically amplifies the contribution of attribute tokens within early BERT layers.

Specifically, the K/V Modulator computes an attribute-conditioned modulation weight that guides the scaling of Key and Value representations for attribute-related tokens.
For each caption, we first build an attribute-conditioned representation that summarizes all attribute prototypes.
Given tokenized caption $T = (t_1, t_2, \ldots, t_L)$ and the extracted attribute span set $\mathcal{S}$, the condition vector $\mathbf{c}$ is constructed as:
\begin{equation}
\mathbf{c}
=
\frac{\sum_{a\in\mathcal{S}} w_a\,\mathbf{p}_a}
{\sum_{a\in\mathcal{S}} w_a},
\end{equation}
where $I_a \subseteq \{1,\ldots,L\}$ denotes the token index set covered by attribute span $a$, 
$w_a = |I_a|$ is its span length, and $\mathbf{p}_a$ denotes the prototype vector of span $a$, computed as the mean embedding of the tokens within that span.

Two modulation functions are then applied to the condition vector to generate near-identity scaling factors for Keys and Values, respectively:
\begin{equation}
\mathbf{s}_K = 1 + \gamma_K \tanh(W^K_2 \, \sigma(W^K_1 \mathbf{c})),
\end{equation}
\begin{equation}
\mathbf{s}_V = 1 + \gamma_V \tanh(W^V_2 \, \sigma(W^V_1 \mathbf{c})),
\end{equation}
where $\mathbf{c} \in \mathbb{R}^{768}$ is the attribute-conditioned vector,
$W^K_1, W^V_1 \in \mathbb{R}^{64\times768}$ and $W^K_2, W^V_2 \in \mathbb{R}^{768\times64}$
denote the projection matrices of the bottleneck MLPs,
$\sigma(\cdot)$ is the GELU activation, and $\tanh(\cdot)$ introduces smooth nonlinearity for bounded modulation.
$\gamma_K$ and $\gamma_V$ are small positive constants (set to 0.1 in our experiments) that control the modulation strength.

At selected early BERT layers (typically the first four), attribute-covered tokens are modulated as:
\begin{equation}
\mathbf{K}^{(\ell)}_t \leftarrow \mathbf{K}^{(\ell)}_t \odot \mathbf{s}_K,\qquad
\mathbf{V}^{(\ell)}_t \leftarrow \mathbf{V}^{(\ell)}_t \odot \mathbf{s}_V,\qquad
t \in \bigcup_{a \in \mathcal{S}} I_a,
\end{equation}
where $\mathbf{K}^{(\ell)}_t$ and $\mathbf{V}^{(\ell)}_t$ are the Key and Value vectors of token $t$ in layer $\ell$, and $\odot$ denotes element-wise multiplication. Tokens outside the extracted attribute spans remain unchanged.

By selectively enhancing Keys, the model increases the attention received by attribute tokens, and by amplifying Values, it strengthens the semantic information written back to contextual representations.  
This dynamic modulation acts as a fine-grained attention bias complementary to Prefix Insertion: the prefix adapter introduces global attribute priors (\emph{what attributes exist}), while KV modulation controls their contribution during contextual reasoning (\emph{how much they matter}).  
Together, they mitigate both pre-encoding bias and intra-encoder competition, effectively activating latent attribute semantics within the frozen text encoder.

\subsection{Training Objective}

DSAA is trained on top of a frozen detector with a three-term objective:
\begin{equation}
\mathcal{L}
= \mathcal{L}_{\text{cls}}
+ \lambda_{\text{det}} \mathcal{L}_{\text{det}}
+ \lambda_{\text{attr}} \mathcal{L}_{\text{attr}},
\end{equation}
where $\mathcal{L}_{\text{cls}}$, $\mathcal{L}_{\text{det}}$, and $\mathcal{L}_{\text{attr}}$ denote the classification, detection, and attribute-aware contrastive losses, respectively. In our experiments, we set $\lambda_{\text{attr}}=0.5$. To improve optimization stability, $\mathcal{L}_{\text{det}}$ is activated only after the attribute-aware branch becomes stable.

\noindent
\textbf{Classification Loss.}
To align image--text pairs globally, we combine binary cross-entropy (BCE) and InfoNCE contrastive losses. Given a batch of $N$ image--text pairs with similarity scores $s_{ij}$ and a temperature parameter $\tau$, the InfoNCE term is formulated as:
\begin{equation}
\mathcal{L}_{\text{InfoNCE}}
= -\frac{1}{N}\!
\sum_{i=1}^{N}
\log
\frac{e^{s_{ii} / \tau}}
{\sum_{j=1}^{N} e^{s_{ij} / \tau}}.
\end{equation}
The total classification loss is
\begin{equation}
\mathcal{L}_{\text{cls}} 
= \mathcal{L}_{\text{BCE}} + \alpha_{\text{nce}} \mathcal{L}_{\text{InfoNCE}},
\end{equation}
where $\alpha_{\text{nce}}$ controls the relative strength of the contrastive term.

\noindent
\textbf{Detection Loss.}
We adopt the standard Grounding DINO detection objective:
\begin{equation}
\mathcal{L}_{\text{det}} =
\mathcal{L}_{\text{loc}} + \mathcal{L}_{\text{cls-det}},
\end{equation}
where $\mathcal{L}_{\text{loc}}$ is the $\ell_1$ box regression loss and $\mathcal{L}_{\text{cls-det}}$ is the classification term under Hungarian matching. Gradients are only propagated through DSAA modules on the text side, while the detector backbone remains frozen.

\noindent
\textbf{Attribute-aware Contrastive Loss.}
Fine-grained detection requires distinguishing instances that share the same category but differ in attributes (e.g., \emph{``red car''} vs. \emph{``blue car''}). 
To explicitly enhance this attribute-level discriminability, 
we introduce an attribute-aware contrastive loss that applies InfoNCE supervision to attribute representations. 
Unlike the global alignment objective in classification, 
this loss focuses on \emph{intra-category} fine-grained separation by enforcing larger margins between different attributes within the same class.

Formally, given $M$ attribute samples, 
we compute the mean attribute logit of the positive caption as $s^{+}_m$ 
and the mean logits of $K_m$ attribute-replacement negatives (belonging to the same category but with different attributes) as $\{s^{-}_{m,1}, s^{-}_{m,2}, \ldots, s^{-}_{m,K_m}\}$. 
The attribute contrastive loss is then formulated as an InfoNCE objective:
\begin{equation}
\mathcal{L}_{\text{attr}} =
- \frac{1}{M}\!
\sum_{m=1}^{M}
\log
\frac{
e^{s^{+}_m / \tau}
}{
e^{s^{+}_m / \tau}
+ \sum_{k=1}^{K_m} e^{s^{-}_{m,k} / \tau}
},
\end{equation}
where $s^{+}_m$ denotes the average attribute logit of the positive sample, 
$s^{-}_{m,k}$ represents that of the $k$-th negative sample within the same category, 
and $\tau$ is a temperature parameter controlling the sharpness of the contrast (set to 0.1 in our experiments). 
This formulation encourages higher scores for matching attribute--region pairs and suppresses incorrect ones, 
thus focusing the contrastive supervision on fine-grained attribute distinctions rather than coarse category differences.

By encouraging tighter intra-attribute clustering and larger inter-attribute margins, 
this loss enhances attribute discriminability and improves the reliability of OVD models in fine-grained detection scenarios.

\section{Experiment}

\subsection{Experimental Settings}

\begin{table*}[t] 
\centering
\caption{\textbf{Main results on the FG-OVD benchmark}. We compare baseline models, their HA-FGOVD-enhanced variants, and our proposed DSAA-enhanced variants across multiple detectors. DSAA consistently achieves superior performance across all fine-grained attributes and difficulty levels.}
\resizebox{\textwidth}{!}{
\begin{tabular}{l|cccccccc|c}
\toprule
Detector & Hard & Medium & Easy & Trivial & Color & Material & Pattern & Transp. & Average \\
\midrule
OWL-ViT(B/16) & 26.2 & 39.8 & 38.4 & 53.9 & 45.3 & 37.3 & 26.6 & 34.1 & 37.7 \\
OWL-ViT(L/14) & 26.5 & 39.3 & 44.0 & 65.1 & 43.8 & 44.9 & 36.0 & 29.2 & 41.1 \\
OWLv2(B/16)  & 25.3 & 38.5 & 40.0 & 52.9 & 45.1 & 33.5 & 19.2 & 28.5 & 35.4 \\
OWLv2(L/14)  & 25.4 & 41.2 & 42.8 & 63.2 & 53.3 & 36.9 & 23.3 & 12.2 & 37.3 \\
Detic         & 11.5 & 18.6 & 18.8 & 69.7 & 21.5 & 38.8 & 30.3 & 24.8 & 29.3 \\
ViLD          & 22.1 & 31.6 & 39.9 & 56.6 & 43.2 & 34.9 & 24.5 & 30.1 & 35.9 \\
CORA          & 13.8 & 20.0 & 20.4 & 35.1 & 25.0 & 19.3 & 22.0 & 27.9 & 22.9 \\
\midrule
GLIP & 37.6 & 41.2 & 39.9 & 56.9 & 45.6 & 37.6 & 38.8 & 35.0 & 41.5 \\
\multicolumn{1}{@{\hspace{4mm}}l|}{+ HA-FGOVD} & 40.2 {\color{red}(+2.6)} & 45.5 {\color{red}(+4.3)} & 43.5 {\color{red}(+3.6)} & 57.1 {\color{red}(+0.2)} & 46.7 {\color{red}(+1.1)} & 40.9 {\color{red}(+3.3)} & 40.7 {\color{red}(+1.9)} & 38.8 {\color{red}(+3.8)} & 44.1 {\color{red}(+2.6)} \\
\multicolumn{1}{@{\hspace{4mm}}l|}{+ \textbf{DSAA (ours)}} & 47.1 {\color{red}(+9.5)} & 52.6 {\color{red}(+11.4)} & 51.1 {\color{red}(+11.2)} & 57.2 {\color{red}(+0.3)} & 52.2 {\color{red}(+6.6)} & 45.1 {\color{red}(+7.5)} & \textbf{48.6} {\color{red}(+9.8)} & 55.3 {\color{red}(+20.3)} & 51.1 {\color{red}(+9.6)} \\
\midrule
DesCo & 43.7 & 40.0 & 34.3 & 61.9 & 46.7 & 32.1 & 19.7 & 43.3 & 40.2 \\
\multicolumn{1}{@{\hspace{4mm}}l|}{+ HA-FGOVD} & 45.4 {\color{red}(+1.7)} & 44.5 {\color{red}(+4.5)} & 38.5 {\color{red}(+4.2)} & 61.8 {\color{green}(-0.1)} & 47.5 {\color{red}(+0.8)} & 35.9 {\color{red}(+3.8)} & 21.9 {\color{red}(+2.2)} & 46.0 {\color{red}(+2.7)} & 42.6 {\color{red}(+2.4)} \\
\multicolumn{1}{@{\hspace{4mm}}l|}{+ \textbf{DSAA (ours)}} & \textbf{52.9} {\color{red}(+9.2)} & \textbf{53.3} {\color{red}(+13.3)} & 46.9 {\color{red}(+12.6)} & 62.4 {\color{red}(+0.5)} & 52.4 {\color{red}(+5.7)} & 47.4 {\color{red}(+15.3)} & 36.1 {\color{red}(+16.4)} & 56.9 {\color{red}(+13.6)} & 51.0 {\color{red}(+10.8)} \\
\midrule
Grounding DINO & 17.0 & 28.4 & 31.0 & 62.5 & 41.4 & 30.3 & 31.0 & 26.2 & 33.5 \\
\multicolumn{1}{@{\hspace{4mm}}l|}{+ HA-FGOVD} & 19.2 {\color{red}(+2.2)} & 32.3 {\color{red}(+3.9)} & 34.0 {\color{red}(+3.0)} & 62.2 {\color{green}(-0.3)} & 41.5 {\color{red}(+0.1)} & 33.0 {\color{red}(+2.7)} & 32.1 {\color{red}(+1.1)} & 29.2 {\color{red}(+3.0)} & 35.4 {\color{red}(+1.9)} \\
\multicolumn{1}{@{\hspace{4mm}}l|}{+ \textbf{DSAA (ours)}} & 38.5 {\color{red}(+21.5)} & 53.0 {\color{red}(+24.6)} & \textbf{59.2} {\color{red}(+28.2)} & \textbf{63.3} {\color{red}(+0.8)} & \textbf{56.0} {\color{red}(+14.6)} & \textbf{52.3} {\color{red}(+21.0)} & 46.5 {\color{red}(+15.5)} & \textbf{63.0} {\color{red}(+36.8)} & \textbf{54.0} {\color{red}(+20.5)} \\
\bottomrule
\end{tabular}
}
\label{tab:fgovd_main_full}
\end{table*}

\noindent \textbf{Dataset.}
The Fine-Grained Open-Vocabulary Detection (FG-OVD) dataset~\cite{bianchi2024devil} is designed to comprehensively evaluate models’ fine-grained attribute discrimination ability.
Each annotation is paired with one positive caption and up to ten hard negatives generated by substituting attribute words while preserving the original sentence structure.
The dataset is divided into four difficulty levels (\textit{Trivial}, \textit{Easy}, \textit{Medium}, \textit{Hard}) and four attribute-focused subsets (\textit{Color}, \textit{Material}, \textit{Pattern}, and \textit{Transparency}).
We train DSAA using the officially released training set of FG-OVD and select only the \textit{hard} subset, which poses the most challenging attribute–category compositions.
We reorganize this subset into 23,432 training instances, each comprising one positive and three hard-negative captions. Despite using a limited amount of training data, DSAA effectively enhances the fine-grained detection capability of OVD models by explicitly strengthening attribute semantics during training.

\noindent \textbf{Evaluation Protocol.}
We follow the official evaluation protocol of FG-OVD \cite{bianchi2024devil}.
Each image is paired with one \textit{positive} caption describing the correct attribute–category composition and several \textit{negative} captions referring to visually or semantically similar distractors.
For each image, the detector predicts region proposals and computes region–text similarity scores with all candidate captions; the caption with the highest similarity is assigned as the region label.
A class-agnostic Non-Maximum Suppression (NMS) is applied to eliminate duplicate detections.
Performance is reported as COCO-style mean Average Precision (mAP) averaged over IoU thresholds from 0.5 to 0.95, evaluated separately on the eight FG-OVD subsets.

\begin{table*}[t]
\centering
\caption{
\textbf{Ablation study on FG-OVD.}
Each column shows mAP on a specific subset, with the change relative to the baseline in parentheses. 
Red indicates improvement and green indicates degradation.}
\resizebox{\textwidth}{!}{
\begin{tabular}{l|ccccccccc}
\toprule
\textbf{Configuration} & Hard & Medium & Easy & Trivial & Color & Material & Pattern & Transp. & Average \\
\midrule
Baseline & 17.0 & 28.4 & 31.0 & 62.5 & 41.4 & 30.3 & 31.0 & 26.2 & 33.5 \\
\midrule
Finetune & 16.6 \textcolor{green}{(-0.4)} & 28.0 \textcolor{green}{(-0.4)} & 24.3 \textcolor{green}{(-6.7)} & 73.6 \textcolor{red}{(+11.1)} & 42.7 \textcolor{red}{(+1.3)} & 29.4 \textcolor{green}{(-0.8)} & 29.3 \textcolor{green}{(-1.7)} & 28.4 \textcolor{red}{(+2.2)} & 34.0 \textcolor{red}{(+0.5)} \\
Learnable Prompt (no APA) & 8.2 \textcolor{green}{(-8.8)} & 10.8 \textcolor{green}{(-17.6)} & 10.2 \textcolor{green}{(-20.8)} & 30.0 \textcolor{green}{(-32.5)} & 18.0 \textcolor{green}{(-23.4)} & 17.8 \textcolor{green}{(-12.5)} & 18.8 \textcolor{green}{(-12.2)} & 17.9 \textcolor{green}{(-8.3)} & 16.5 \textcolor{green}{(-17.0)} \\
\midrule
APA & 35.7 \textcolor{red}{(+18.7)} & 48.8 \textcolor{red}{(+20.4)} & 51.4 \textcolor{red}{(+20.4)} & 60.0 \textcolor{green}{(-2.5)} & 52.0 \textcolor{red}{(+10.6)} & 45.7 \textcolor{red}{(+15.4)} & 44.7 \textcolor{red}{(+13.7)} & 58.8 \textcolor{red}{(+32.6)} & 49.6 \textcolor{red}{(+16.1)} \\
APA + AttrLoss & 37.1 \textcolor{red}{(+20.1)} & 53.0 \textcolor{red}{(+24.6)} & 55.3 \textcolor{red}{(+24.3)} & 60.6 \textcolor{green}{(-1.9)} & 56.3 \textcolor{red}{(+14.9)} & 47.2 \textcolor{red}{(+16.9)} & 39.9 \textcolor{red}{(+8.9)} & 62.3 \textcolor{red}{(+36.1)} & 51.5 \textcolor{red}{(+18.0)} \\
APA +  AttrLoss + K/V Modulator & \textbf{38.5} \textcolor{red}{(+21.5)} & \textbf{53.0} \textcolor{red}{(+24.6)} & \textbf{59.2} \textcolor{red}{(+28.2)} & \textbf{63.3} \textcolor{red}{(+0.8)} & \textbf{56.0} \textcolor{red}{(+14.6)} & \textbf{52.3} \textcolor{red}{(+21.0)} & \textbf{46.5} \textcolor{red}{(+15.5)} & \textbf{63.0} \textcolor{red}{(+36.8)} & \textbf{54.0} \textcolor{red}{(+20.5)} \\
\bottomrule
\end{tabular}
}
\label{tab:ablation_fgovd}
\end{table*}

\begin{figure*}[t]
    \centering
    \includegraphics[width=\textwidth]{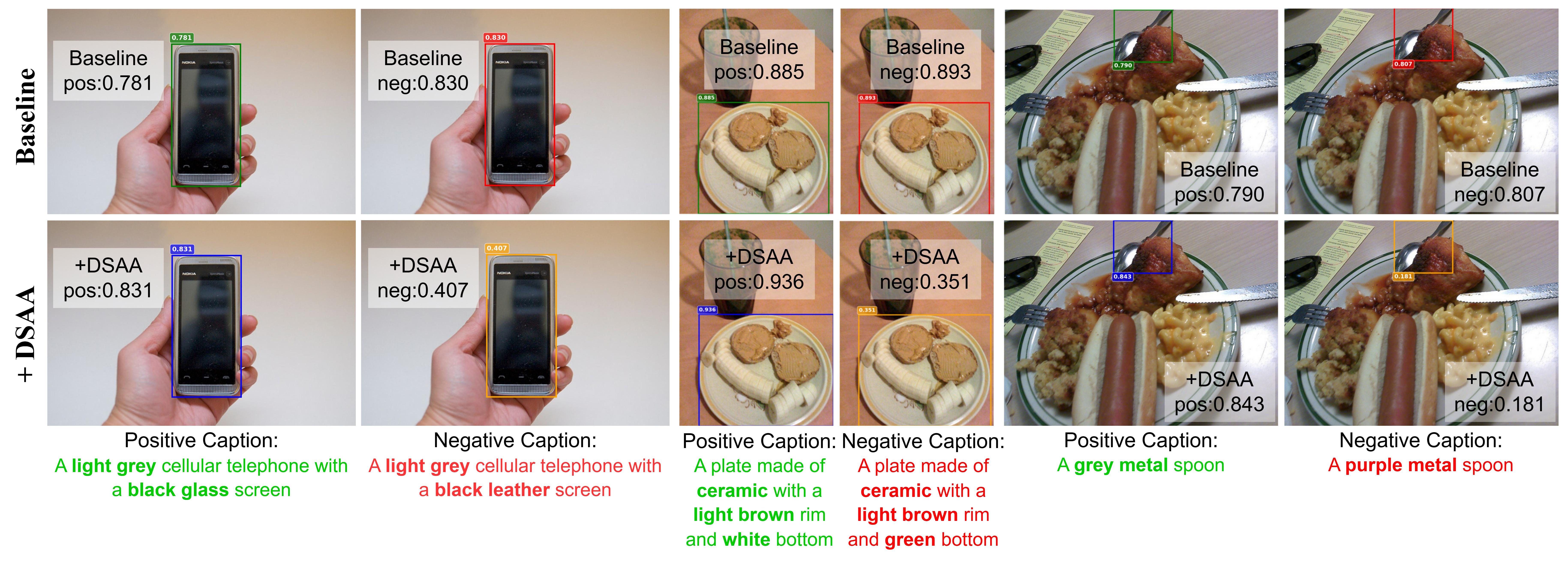}
    \caption{
    \textbf{Qualitative results of Grounding DINO with and without DSAA on the FG-OVD benchmark.}
    Each row presents detection results under attribute-rich text queries.
    The top row shows predictions from the baseline, and the bottom row shows those from DSAA.
    Green/blue boxes denote positive captions (correct attribute–object matches), while red/orange boxes denote negative captions (incorrect or mismatched compositions).
    The gray boxes display the model’s confidence scores for each queried caption.}
    \label{fig:qualitative_results}
\end{figure*}

\begin{table}[t]
\centering
\caption{\textbf{Inference latency on Grounding DINO with different LLMs.}}
\label{tab:llm_latency}
\vspace{3pt}
\small
\setlength{\tabcolsep}{3pt}
\begin{tabular}{p{3.0cm}ccc}
\toprule
Method & LLM (ms) & Detector (ms) & Overall (ms) \\
\midrule
Baseline & --    & 345.26 & 345.26 \\
DSAA + Qwen3-4B           & 19.93 & 354.29 & 374.22 \\
DSAA + Qwen3-8B           & 42.48 & 354.29 & 396.77 \\
\bottomrule
\end{tabular}
\end{table}
\noindent \textbf{Implementation Details.}
All experiments are conducted on four NVIDIA RTX 3090 GPUs. 
The model is trained for 4 epochs using the AdamW optimizer with a learning rate of $1\times10^{-4}$, weight decay of 0.01, and a batch size of 3 per GPU. We further analyze the runtime of the LLM extractors and the detector,  as summarized in Table~\ref{tab:llm_latency}.
Both Qwen3-8B and Qwen3-4B are served with vLLM, and all latency
measurements are conducted on a single NVIDIA RTX~3090 GPU. Unless otherwise specified, all main experiments in this paper use Qwen3-8B as the default LLM extractor.

\subsection{Experimental Results}

We integrate DSAA into three representative open-vocabulary detection models, including Grounding DINO, GLIP, and DesCo, and compare them with established baselines such as OWL-ViT, Detic, ViLD, CORA. 
We present the main experimental results of the proposed DSAA framework on the FG-OVD benchmark in \cref{tab:fgovd_main_full}.  
Although Grounding DINO excels in open-vocabulary localization, it struggles with fine-grained detection.
This limitation arises because category tokens tend to dominate the region--text alignment process, causing attribute semantics to be weakened in the text encoder. To address this, DSAA explicitly introduces attribute priors via the Attribute Prefix Adapter and preserves attribute semantics in early transformer layers through the K/V Modulator. Consequently, DSAA restores fine-grained attribute–object binding, yielding a substantial +20.5 average mAP improvement in Grounding DINO.
Furthermore, models like GLIP and DesCo exhibit stronger initial fine-grained alignment due to their caption-level or description-rich supervision. However, their pre-training primarily encourages holistic sentence-level representations, often underrepresenting explicit attribute semantics crucial for fine-grained discrimination.  Despite this, when applied to GLIP and DesCo, DSAA also produces significant improvements of +9.6 mAP and +10.8 mAP, respectively. These consistent gains demonstrate DSAA's effectiveness in further refining attribute understanding, even for models already possessing some fine-grained capabilities. Notably, DSAA is a conditional and non-invasive framework: when no attribute tokens are extracted from the text prompt, inference naturally falls back to the original detector. Accordingly, DSAA preserves standard open-vocabulary detection performance on general benchmarks, matching the baseline Grounding DINO on COCO and LVIS (48.4 mAP and 27.4 mAP, respectively), while significantly improving fine-grained detection.

\subsection{Ablation Study}
We conduct ablations on Grounding DINO to evaluate the contribution of each component in DSAA, with the results summarized in \cref{tab:ablation_fgovd}.

\noindent \textbf{Learnable Prompt (no APA).}
Before introducing the Attribute Prefix Adapter, we first examine a variant that replaces the adapter with a fixed learnable prompt, following the conventional prompt-tuning paradigm.  
In this setup, we use four learnable prefixes to evaluate the impact of static prompts.  
As shown in \cref{tab:ablation_fgovd}, this configuration results in a noticeable performance drop compared to the baseline, suggesting that static prompts fail to capture the diversity of attribute semantics, leading to reduced attribute separability.  

\noindent \textbf{Attribute Prefix Adapter.}
The APA injects attribute-specific prefixes into the embedding sequence, providing explicit semantic anchors for the frozen encoder.  
This adaptive mechanism allows the model to better capture and represent fine-grained attribute cues that are otherwise suppressed by category tokens.  
As shown in \cref{tab:ablation_fgovd}, APA alone achieves a substantial improvement over the baseline, increasing the average mAP from 33.5 to 49.6 (+16.1).  

\noindent \textbf{Attribute-Aware Contrastive Loss.}
Adding the attribute-aware contrastive loss on top of APA further improves the average mAP from 49.6 to 51.5 (+1.9). This loss explicitly encourages separation between same-category instances with different attributes, which helps the model learn more discriminative attribute-sensitive representations. The gain indicates that, beyond input-level activation, explicit fine-grained supervision is also beneficial for stabilizing attribute discrimination.

\noindent \textbf{K/V Modulator.}
Finally, introducing the K/V Modulator on top of APA and the attribute-aware contrastive loss further improves the average mAP from 51.5 to 54.0 (+2.5), yielding the best overall performance. With APA already providing explicit attribute anchors and AttrLoss enforcing fine-grained separation, the K/V Modulator can more effectively amplify attribute-relevant signals during contextual encoding. This complementary improvement shows that preserving and strengthening attribute semantics inside early encoder layers is important in addition to activating them at the embedding stage.

\subsection{Qualitative Results}
\noindent\textbf{Qualitative Comparison.}
We visualize predictions using attribute-rich queries from the FG-OVD dataset to evaluate DSAA’s impact on fine-grained grounding (\cref{fig:qualitative_results}).  
Compared to the baseline, DSAA improves attribute-object grounding accuracy and reduces common failures, such as mismatched color or material assignments.  
DSAA assigns higher confidence to correct attribute-object pairs and suppresses mismatched pairs, indicating enhanced sensitivity to subtle visual variations within the same category.  
In contrast, the baseline often misidentifies positive and negative samples, showing an overreliance on category semantics rather than attribute cues.

\begin{figure}[t]
  \centering
  \captionsetup{skip=3pt}
  \captionsetup[subfigure]{justification=centering,singlelinecheck=false}

  \begin{subfigure}[t]{0.49\linewidth}
    \centering
    \includegraphics[width=\linewidth]{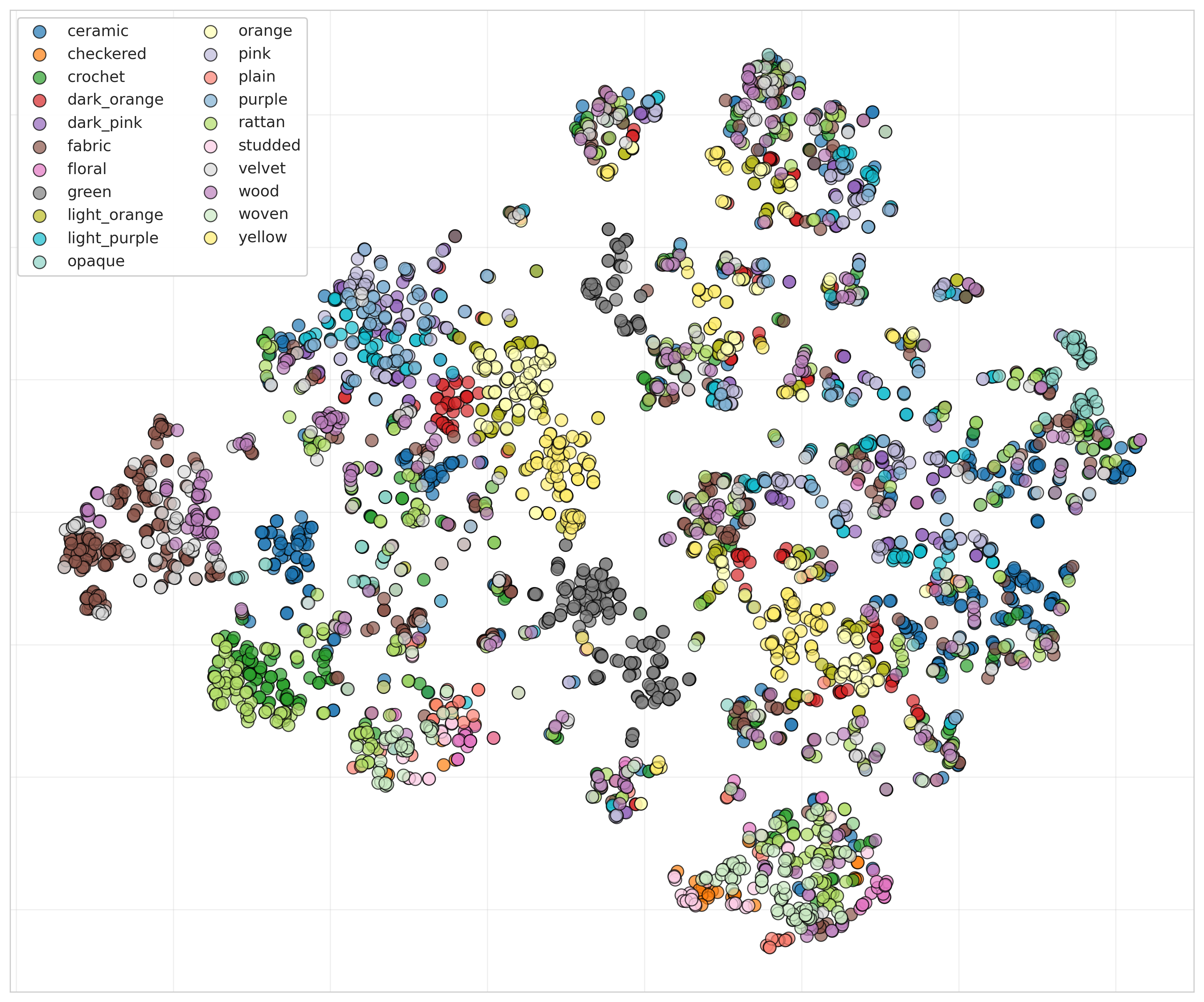}
    \caption{Baseline}
    \label{fig:tsne_base}
  \end{subfigure}
  \hfill
  \begin{subfigure}[t]{0.49\linewidth}
    \centering
    \includegraphics[width=\linewidth]{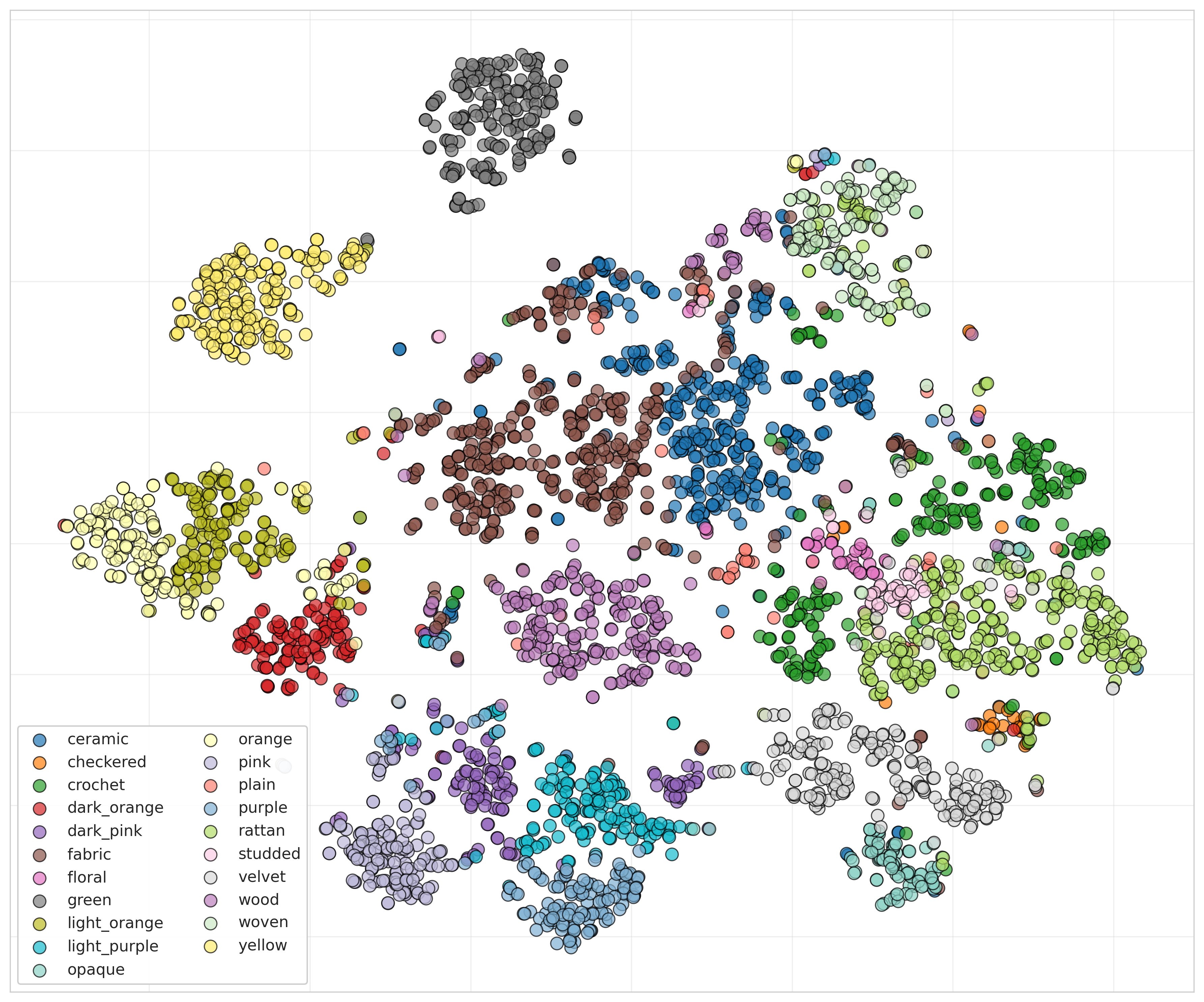}
    \caption{DSAA}
    \label{fig:tsne_dsaa}
  \end{subfigure}

  \caption{\textbf{t-SNE visualization of attribute embeddings on Grounding DINO.}
  DSAA forms more compact and semantically consistent clusters across attributes, demonstrating its effectiveness in improving the separation and coherence of attribute representations compared to the baseline.}
  \label{fig:cluster_visualization}
\end{figure}

\begin{figure}[t]
    \centering
    \includegraphics[width=0.80\linewidth]{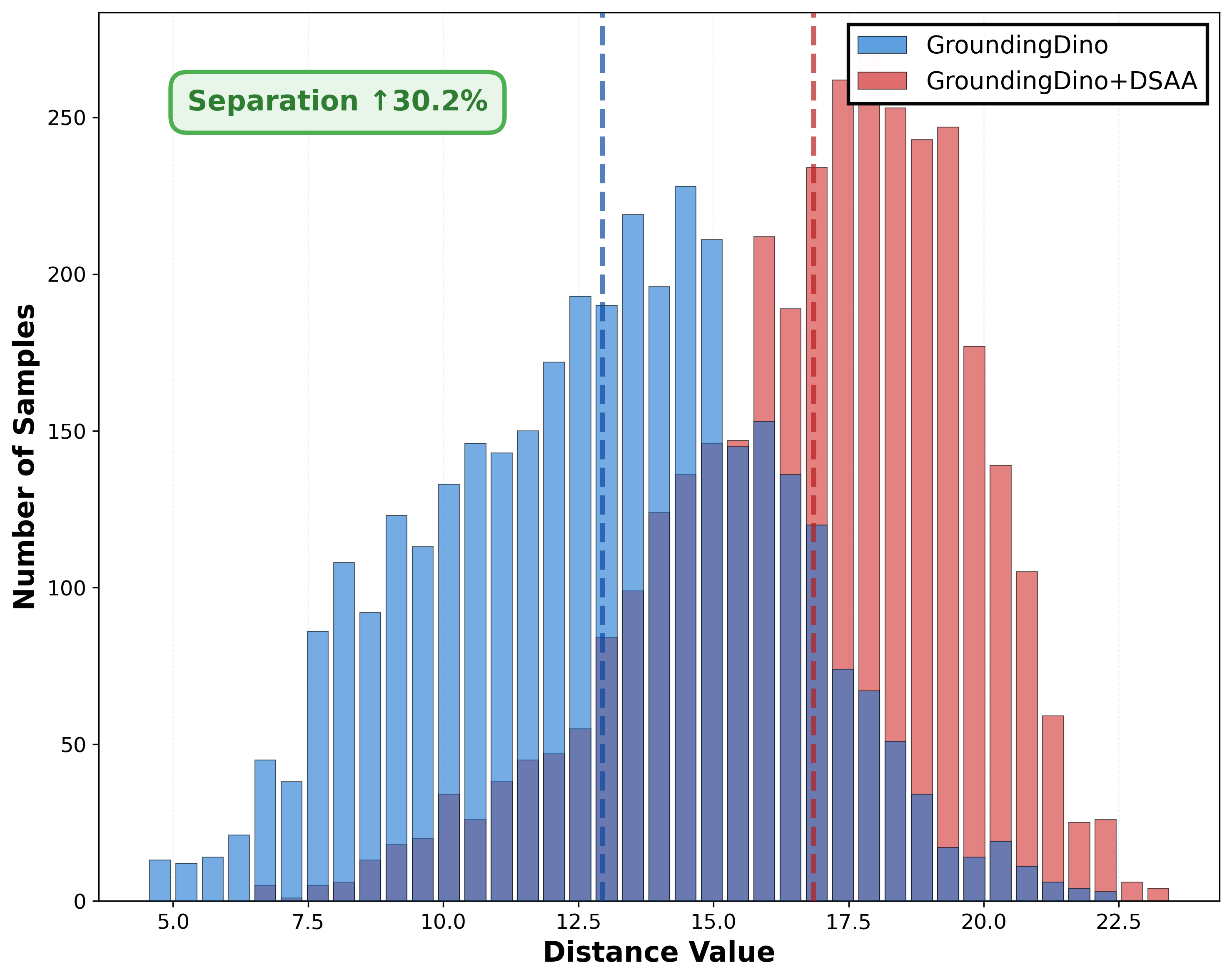}
    \caption{\textbf{Feature distance distribution.} DSAA increases the separation between positive and negative samples by 30.2\%, demonstrating enhanced feature discriminability.}
    \label{fig:distance_distribution}
\end{figure}

\begin{table}[t]
\caption{\textbf{Quantitative evaluation of category suppression.}
DSAA alleviates the marginalization of attributes under category dominance, achieving higher average cosine distances and thus stronger attribute separability compared to the baseline.}
  \centering
  \small
  \setlength{\tabcolsep}{8pt}      
  \renewcommand{\arraystretch}{1.15} 
  \begin{tabular}{lcc}
    \toprule
    \textbf{Category Type} & \textbf{Baseline} & \textbf{DSAA} \\
    \midrule
    Neutral Category & 0.2302 & \textbf{0.3401} \\
    Explicit Category& 0.1347 & \textbf{0.3360} \\
    \bottomrule
  \end{tabular}
  \label{tab:dsaa_suppression}
\end{table}
\noindent\textbf{Attribute Representation Analysis.}  
We further analyze DSAA’s ability to mitigate \textit{category dominance}, where strong category semantics suppress attribute representation.  
When a prompt contains an explicit category (e.g., ``green dog''), attribute cues (e.g., ``green'' vs.\ ``black'') become less distinguishable compared to neutral prompts (e.g., ``green object'').  
To quantify this, we construct two groups of prompts: Neutral Category and Explicit Category.  
The Neutral Category group uses general category terms such as \textit{object}, \textit{item}, or \textit{thing}, which carry minimal semantic bias.  
In contrast, the Explicit Category group uses specific category nouns like \textit{dog}, \textit{chair}, or \textit{plate}, which introduce stronger semantic cues and may dominate attribute representations.
As shown in \cref{tab:dsaa_suppression}, the baseline exhibits clear suppression under explicit categories, whereas DSAA effectively alleviates this effect and improves attribute separability.
Feature analyses further validate this trend.  
The t-SNE visualization (\cref{fig:cluster_visualization}) shows that DSAA produces more compact and semantically consistent attribute clusters, indicating better preservation of attribute semantics.  
Meanwhile, the feature distance distribution (\cref{fig:distance_distribution}) demonstrates that DSAA enlarges the separation between positive and negative samples by 30.2\%, confirming its enhanced discriminability and fine-grained differentiation capability.

\section{Conclusion}
In this work, we identified a core limitation of current open-vocabulary detectors: attribute semantics are marginalized under strong category priors, hindering fine-grained recognition. We addressed this issue with Dual-Stage Attribute Activation (DSAA), a non-invasive framework that explicitly activates attribute information within frozen text encoders. DSAA substantially improves fine-grained performance, achieving +20.5 average mAP on the FG-OVD benchmark when integrated with Grounding DINO, while preserving standard category detection capability when attribute cues are absent. These results demonstrate that fine-grained knowledge embedded in pre-trained models can be effectively unlocked without large-scale retraining. Beyond OVD, the principles behind DSAA offer a promising direction for enhancing attribute sensitivity in broader vision–language systems.
\section*{Acknowledgments}

This work was supported in part by the Beijing Natural Science Foundation (No.4262060), 
the Opening Project of the State Key Laboratory of General Artificial Intelligence (BIGAI/Peking University, Project No. SKLAGI2025OP23), 
the National Key R\&D Program of China (No. 2021ZD0109802), 
and the High-performance Computing Platform of BUPT.

{
    \small
    \bibliographystyle{ieeenat_fullname}
    \bibliography{main}
}

\clearpage  
\end{document}